\documentclass{article}

\usepackage{microtype}
\usepackage{graphicx}
\usepackage{subcaption}
\usepackage{booktabs} 
\usepackage{makecell}

\usepackage{hyperref}

\usepackage[preprint]{icml2026}

\usepackage{amsmath}
\usepackage{amssymb}
\usepackage{mathtools}
\usepackage{amsthm}
\usepackage{xspace}
\usepackage{placeins}

\usepackage[capitalize,noabbrev]{cleveref}

\theoremstyle{plain}

\theoremstyle{definition}

\theoremstyle{remark}

\usepackage[textsize=tiny]{todonotes}

\icmltitlerunning{Self-supervised multi-modal spatial transcriptomics learning}

\newcommand{\modelname}{\xspace\texttt{HistoTx}\xspace}

\begin{document}

\twocolumn[
  \icmltitle{\modelname: Early fusion of H\&E images and spatial transcriptomics at varying spatial transcriptomics resolution with self-supervised learning}



  \icmlsetsymbol{equal}{*}

  \begin{icmlauthorlist}
    \icmlauthor{Alice Driessen}{equal,kaiko}
    \icmlauthor{Emil Malmsten}{kaiko}
    \icmlauthor{Steffen Wolf}{kaiko}
    \icmlauthor{Bernd Illing}{CHUV,SIB}
    \icmlauthor{Isinsu Katircioglu}{CHUV,SIB}
    \icmlauthor{Adriano Martinelli}{CHUV,SIB}
    \icmlauthor{Rolf Stahel}{ETOP}
    \icmlauthor{Solange Peters}{CHUV}
    \icmlauthor{Enriqueta Felip}{VHUH}
    \icmlauthor{Sanjay Popat}{TRM}
    \icmlauthor{Krisztian Homicsko}{ONCCHUV}
    \icmlauthor{Raphael Gottardo}{CHUV,SIB,UNIL,EPFL}
    \icmlauthor{Marianna Rapsomaniki}{CHUV,SIB,UNIL}
    \icmlauthor{Fei Tang}{kaiko}
  \end{icmlauthorlist}

  \icmlaffiliation{kaiko}{kaiko.ai, Zürich, Switzerland}
  \icmlaffiliation{UNIL}{Faculty of Biology and Medicine, University of Lausanne, Lausanne, Switzerland}
  \icmlaffiliation{CHUV}{Biomedical Data Science Center, Lausanne University Hospital, Lausanne, Switzerland}
  \icmlaffiliation{SIB}{Swiss Institute of Bioinformatics (SIB), Lausanne, Switzerland}
  \icmlaffiliation{ONCCHUV}{Department of Oncology, Lausanne University Hospital, Lausanne, Switzerland}
  \icmlaffiliation{EPFL}{School of Life Sciences, Ecole Polytechnique Fédérale de Lausanne, Lausanne, Switzerland}
  \icmlaffiliation{ETOP}{ETOP IBCSG Partners Foundation}
  \icmlaffiliation{VHUH}{Vall d'Hebron University Hospital, Barcelona, Spain}
  \icmlaffiliation{TRM}{The Royal Marsden, Londen, United Kingdom}
  \icmlcorrespondingauthor{Alice Driessen}{alice.driessen@kaiko.ai}
  \icmlcorrespondingauthor{Fei Tang}{fei@kaiko.ai}

  \icmlkeywords{Machine Learning, ICML}

  \vskip 0.3in
]



\printAffiliationsAndNotice{}  

\begin{abstract}
Spatial transcriptomics maps gene expression across tissue sections and, when integrated with whole slide images, jointly captures both morphological architecture and molecular activity to provide a rich, multimodal view of tissue biology. ML approaches combining spatial transcriptomics and WSIs typically learn a mapping from morphological features to molecular profiles via contrastive alignment or supervised fine-tuning of pathology foundation models. In both cases, transcriptomics serves as a training signal rather than a true input modality, leaving inference fundamentally image-driven. We propose an early-fusion vision transformer architecture in which transcript tokens are constructed at arbitrary spatial resolution, merged with image patch tokens, and jointly processed through a shared transformer stack for deep cross-modal interaction. Starting from a pretrained vision-only pathology foundation model, we train \modelname by continuing self-supervised pretraining on paired image and spatial transcriptomics data and demonstrate that: 1) at inference with image only, performance remains on par with the baseline on most tasks while improving on tasks that inherently benefit from molecular knowledge, such as gene expression prediction; 2) when both modalities are available at inference, jointly providing image and transcript inputs outperforms either modality alone.

\end{abstract}

\section{Introduction}
Hematoxylin and eosin (H\&E)-stained whole slide images (WSIs) are the gold standard for cancer diagnosis, routinely used for tumor detection, subtyping, and clinical decision making, yet their interpretation remains a labor-intensive task demanding years of specialized pathologist training \cite{macbeth_pathology_2015}. Recent advances in artificial intelligence have led to the emergence of large pathology foundation models trained on vast collections of WSIs, achieving strong performance across a broad range of clinically relevant tasks \cite{wang_transformer-based_2022, chen_towards_2024, xu_whole-slide_2024}, yet their capabilities remain limited for tasks that could benefit from additional molecular knowledge. 

In parallel, advances in spatial transcriptomics (ST) technologies now enable molecular profiling directly on tissue sections, facilitating precise cell type identification, detection of functional cell states, and richer tumor characterization \cite{jain_spatial_2024}. ST data comes in different formats depending on the platform. For example, high-resolution Xenium data consists of individual transcript detections, each annotated with its spatial location and gene identity but limited to a fixed gene panel. By contrast, Visium data is whole-transcriptome based and consists of a regular grid of barcoded spots, each with a diameter of 55 µm, where transcripts are aggregated into a single expression profile per spot. Due to the different nature and granularity of the ST data it is nontrivial to train a model on both Xenium and Visium data. In fact most models use only on spot-based \cite{chen_visualomics_2025, chung_accurate_2024, pang_leveraging_2021, xie_spatially_nodate, hemker_towards_2026, zhan_sage-fm_nodate} or probe-based data \cite{fu_spatial_2025, martinelli_learning_2026, gindra_large-scale_2025} for training. To our best knowledge, only MINT \cite{lee_mint_2026} is able to encode both modalities with one model, however this model is limited to one learnable ST token.

To leverage molecular information alongside histology, early approaches trained supervised models to predict gene expression directly from H\&E images \cite{schmauch_deep_2025, chung_accurate_2024, nonchev_deepspot_2025, pang_leveraging_2021, hu_histoprism_2026}. More recently, methods inspired by CLIP have employed contrastive learning to align image and ST embeddings in a shared latent space, using the transcriptomic modality merely as a supervision signal to guide visual representation learning \cite{chen_visualomics_2025, zhang_care_2026, xie_spatially_nodate, hemker_towards_2026}. Multimodal fusion is reduced to late-stage feature-level concatenation or averaging, rather than being deeply integrated into the model architecture. We argue that this image-centric strategy is well motivated in practice: given the high cost of ST, molecular profiles are rarely available in routine clinical workflows, whereas H\&E images are inexpensive and virtually ubiquitous. Leveraging limited ST data as a supervision signal to endow the image encoder with molecular knowledge is therefore a natural and clinically practical design choice. However, existing contrastive approaches apply this supervision crudely, typically at the coarse patch level, failing to exploit the fine-grained spatial correspondence between transcript distributions and image content. At the same time, in the cases where transcriptomics data are available at inference, deeper and more principled multimodal fusion beyond simple concatenation or averaging remains desirable to fully exploit the complementary information carried by both modalities, as can be achieved by early-fusion \cite{martinelli_learning_2026}.

We present \modelname, an early-fusion multimodal transformer building on \cite{martinelli_learning_2026} that integrates H\&E images and ST. We extend the architecture to accommodate both Visium and Xenium platforms by constructing spatially aligned image and transcript tokens at multiple spatial resolutions, merging them into a joint token sequence, and processing them through a shared transformer stack for deep cross-modal interaction. \modelname is trained by continuing DINOv2~\cite{oquab_dinov2_2024} self-supervised pretraining on paired image and ST data, warm-started from a pretrained vision-only pathology foundation model. By constructing tokens at multiple spatial resolutions, the model captures biological structures across scales, from subcellular transcript patterns to tissue-level morphology. The two main contributions of this work are:
\begin{itemize}
    \item \textbf{Multimodal self-supervised pretraining:} a training strategy that continues DINOv2 pretraining on paired image and ST data from a vision-only pathology foundation model, yielding a model that retains strong image-only performance, improves on tasks requiring molecular knowledge, and further benefits when both modalities are available at inference.
    \item \textbf{Flexible transcript tokenization strategy:} This enables a single backbone to handle both Xenium and Visium data, allowing training on the full HEST-1k~\cite{jaume_hest-1k_2024} dataset and broadening exposure to diverse tissues, gene panels, and spatial scales.
\end{itemize}



\begin{figure*}[ht]
  \begin{center}
    \centerline{\includegraphics[width=\textwidth]{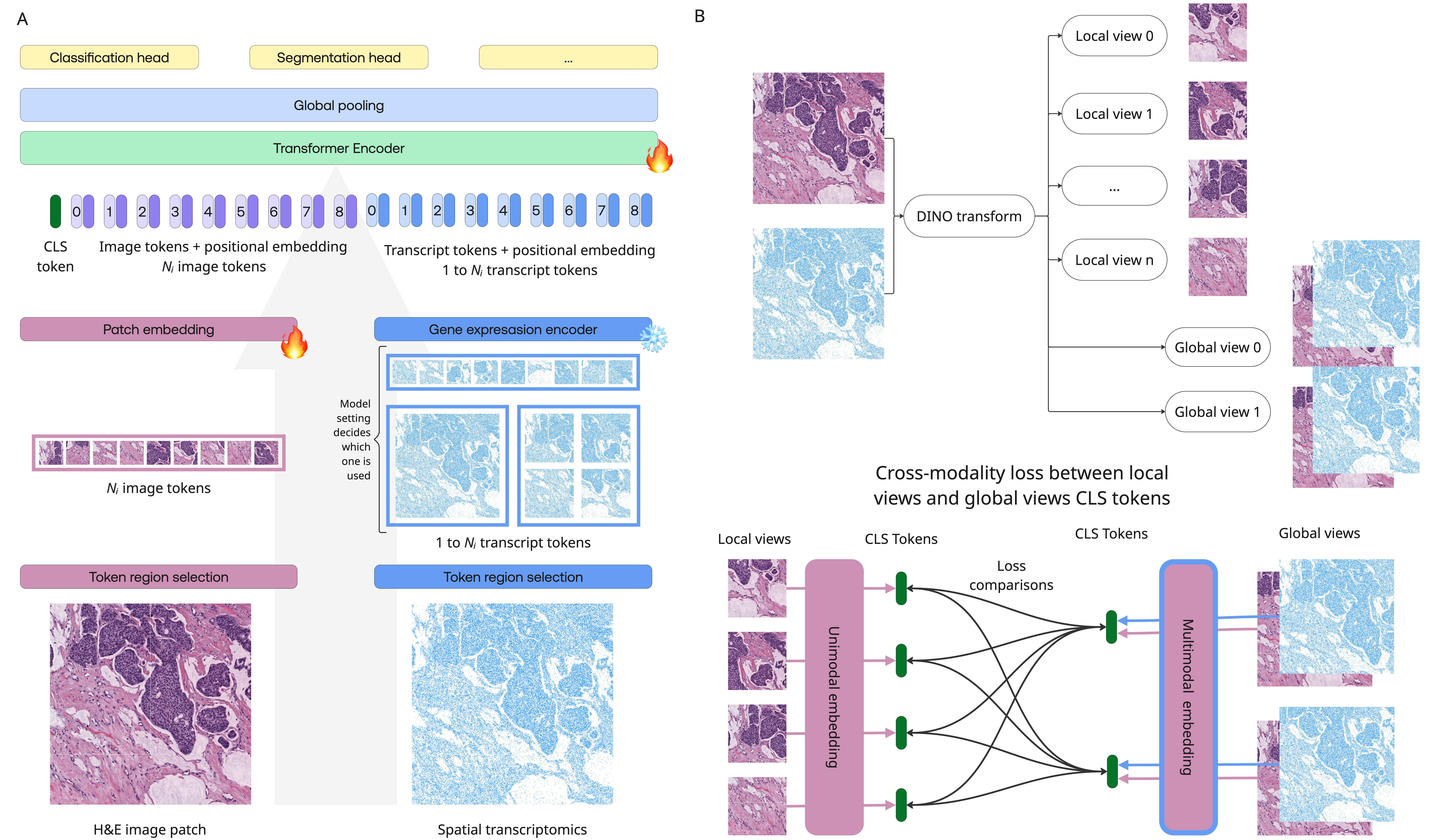}}
    \caption{
      Early-fusion multimodal model architecture and cross-modal loss. A) Input patches contain the H\&E image and the corresponding spatial transcriptomics. Both modalities are tokenized into smaller token regions and embedded with modality-specific token encoders. The tokens of both modalities are concatenated and fed through a ViT block. The snow flake emoji indicates the frozen part of the model and the flame the part where model weights are optimized. B) DINO transforms created $n$ local views and 2 global views. By removing transcripts from the local views, the DINO local view loss on the CLS tokens becomes a cross-modality loss. 
    }
    \label{fig:model_architecture}
  \end{center}
\end{figure*}

\section{Methods}
\subsection{Model architecture and training}
\modelname is a multimodal foundation model that jointly encodes tissue morphology from histology images and gene expression from spatial transcriptomics data (see figure \ref{fig:model_architecture}). It comprises a vision transformer based morphology encoder as the backbone, and a gene expression encoder. 

\subsubsection{Unimodal tokenization}
We train on a mixture of Visium spot-based and probe-based Xenium data. Xenium provides individual transcript coordinates, enabling tokens to be constructed at arbitrary spatial resolution by counting per-gene detections within non-overlapping $T \times T$ pixel regions, each yielding one transcript count vector. By varying the size of $T$, we can adapt the number of transcript tokens. For an image of size $224 \times 224$ pixels, this yields 256, 64, 16, or 1 tokens for $T=14, 28, 56, \text{or}~~ 256$ respectively. The physical area covered by each token depends jointly on $T$ and the image resolution: at $\text{mpp}=0.5$, $T=14$ corresponds to a $7 \times 7 \mu m^2$ region, while at $\text{mpp}=2.0$, $T=56$ covers $112 \times 112\ \mu m^2$. By varying $T$ and mpp, tokens can be constructed at different spatial scales, capturing structures from subcellular to tissue-level.


Since Visium data is spot-based, providing only aggregated expression at fixed spot locations rather than individual transcript coordinates, we follow HEST-1k~\cite{jaume_hest-1k_2024} and extract $224 \times 224$ pixel images at $\text{mpp}=0.5$ centered on each spot yielding one transcript token per image.  At coarser resolutions, a single image encompasses multiple spots, up to 4 and 16 at $\text{mpp}=1$ and $\text{mpp}=2$ respectively, each contributing its aggregated count vector as a separate transcript token.

Transcript count vectors are encoded by a gene expression encoder: here we use the pretrained gene foundation model Geneformer~\cite{theodoris_transfer_2023}, producing transcript token embeddings of dimension $D_{\text{expr}}$. 

Images are tokenized via the patch embedding layer of the morphology encoder, producing $N_I$ image token embeddings of dimension $D_{\text{morph}}$ per image. We deliberately decouple the spatial resolution of transcript tokens from that of image tokens for two reasons: first, high-resolution Xenium data enabling fine-grained transcript tokenization is scarce and decoupling allows Visium data to complement Xenium in training; second, at very high resolution, individual transcript tokens may be dominated by noise rather than meaningful signal.

\subsubsection{Multimodal fusion}
After tokenization, transcript embeddings are projected to match the image token dimension via a linear layer and rescaled to match the norm of the image tokens. The morphology encoder's positional embeddings are then applied to both modalities; when the number of transcript tokens $N_T$ differs from the number of image tokens $N_I$, positional embeddings are adapted to the transcript spatial grid via dynamic spatial resampling \cite{dosovitskiy_image_2021}. The fused sequence $[\text{CLS} \mid \mathbf{z}_I^{1:N_I} \mid \mathbf{z}_G^{1:N_G}]$ of length $1 + N_I + N_G$ is passed through the remaining ViT blocks. The output CLS token serves as the global multimodal representation, while patch and transcript tokens occupy the remaining positions. At inference, the model supports CLS, mean, max, or flatten pooling over the output sequence.



\subsubsection{Self-supervised learning}
Starting from a pathology foundation model pretrained on large WSI collections, we continue pretraining on paired image-transcript data using DINOv2. To accommodate the multimodal setting, we extend the notion of an image view to image-transcript pairs. For Xenium data, the random crop and resize transformation is applied jointly to both modalities, retaining only transcript detections falling within the cropped region; the cropped image then undergoes standard photometric augmentation in DINO transform (colour jitter, Gaussian blur, etc.). To introduce a cross-modality learning signal within the DINOv2 framework, transcripts are included only in global views while local views remain image-only, such that global CLS tokens are multimodal and local CLS tokens are image-only. The standard DINO local crop loss then enforces consistency between these two, driving cross-modal knowledge distillation from transcripts into the visual encoder. (see figure \ref{fig:model_architecture}B). For Visium data, where we always obtain one single transcript token per patch, the token is retained for global views and omitted for local views. 

Our experiments are conducted primarily on a ViT-B/14 backbone initialized from an internal Midnight \cite{karasikov_training_2025} checkpoint of the same size. The gene expression encoder uses the \textit{gf-6L-10M-i2048} variant of GeneFormer, kept frozen throughout training. Full model and training hyperparameters are reported in Appendix \ref{app:model_details}.

\begin{table*}[t]
  \caption{
  Gene expression tasks: Hbench: HEST-Bench, Hexpr: HEST expression prediction data from \cite{jaume_hest-1k_2024}; HexprIO: Hest expression task with image only input. BEAT: ETOP BEAT-meso prediction of 100 held-out genes from image+remaining 380 genes. BEATIO: ETOP BEAT-meso prediction of 100 genes from image only. Classification tasks: C16: Camelyon16, BACH, BRACS, BH: BreakHis, GTMA: Gleason TMA, PR: PANDA Radboud; Segmentation tasks: MNS: MoNuSAC, TTS: Tan Tock Seng, CNP: Consep. See Appendix \ref{app:task_details} for more task details. Supervised model as described in \cite{martinelli_learning_2026}, using vit small patch16 224.augreg in21k as morphology encoder and a trainable/unfrozen MLP as expression encoder.
  MCAcc: Multiclass accuracy, BBAcc: Binary Balance accuracy, MDS: Monai Dice Score, PC: Mean Pearson correlation coefficient.}  
  \label{tab:benchmark_main}
  \begin{center}
    \begin{small}
      \begin{sc}
        \begin{tabular}{ll|llll}
        \toprule

Benchmark   & Metric & Image only   & \modelname    & SEAL+CONCH        & Supervised       \\
\midrule
HBench      & PC    &   0.31            & \textbf{0.38}     & 0.35              & - \\
Image Hexpr & PC    &   -               & \textbf{0.36}     & -                 & - \\
HexprIO     & PC    &   0.26            & \textbf{0.30}     & 0.28              & - \\
BEAT        & PC    &   -               & 0.66              & -                 & \textbf{0.82} \\
BEATIO      & PC    &   0.53            & \textbf{0.58}     & -                 & - \\
C16         & BBAcc &   0.79            & 0.87              & \textbf{0.89}     & - \\
BACH        & MCAcc &   \textbf{0.88}   & 0.82              & 0.82              & - \\
BRACS       & MCAcc &   \textbf{0.64}   & 0.60              & 0.61              & - \\
BH          & MCAcc &   \textbf{0.82}   & 0.73              & 0.68              & - \\
GTMA        & MCAcc &   \textbf{0.77}   & 0.69              & 0.73              & - \\
PR          & MCAcc &   0.61            & \textbf{0.67}     & 0.66              & - \\
MNS         & MDS   &   0.61            & 0.62              & \textbf{0.63}     & - \\
TTS         & MDS   &   0.31            & \textbf{0.37}     & \textbf{0.37}     & - \\
CNP         & MDS   &   0.61            & \textbf{0.62}     & \textbf{0.62}     & - \\

\bottomrule
         \end{tabular}
      \end{sc}
    \end{small}
  \end{center}
  \vskip -0.1in
\end{table*}
\subsection{Training datasets}
We use the \textbf{HEST1k} \cite{jaume_hest-1k_2024} and \textbf{TCGA} \cite{tcga_cancer_2025} dataset to train \modelname. Training details and dataloader weights are listed in Appendix \ref{app:model_details}. TCGA is an image-only dataset used to preserve the model's base image processing capabilities. Since, HEST1k contains H\&E images with spot-based (Visium Spatial Transcriptomics) or Xenium data we use this data for multimodal training, as well as image-only training.

\section{Results}
We trained a family of \modelname models differing in how expression tokens are constructed from Xenium data; implementation details and full results are provided in Appendix \ref{app:model_details}, Tables \ref{tab:benchmark} and \ref{tab:benchmark2}. Here we focus on the variant using one expression token per Visium spot and 16 expression tokens per Xenium image of size 224x224 px. We hypothesize that too many small tokens introduce noise as transcript counts at very fine spatial scales are unreliable, while too few tokens sacrifice spatial resolution and molecular detail.

We compare \modelname against two baselines: the vision-only foundation model used as our starting point, and SEAL~\cite{hemker_towards_2026}, which adapts a pretrained image encoder via CLIP-style multimodal fine-tuning. For SEAL we use the CONCH backbone, as CONCH is also a ViT-B model, comparable in scale to our \modelname backbone. Downstream benchmarks include image-only classification (Camelyon16, Panda Radboud), image-only segmentation (Tan Tock Seng, MoNuSAC), and gene expression prediction in both image-only and multimodal settings (HEST-Bench, HEST-Expr); see Appendix~\ref{app:task_details} for task descriptions. 

\modelname outperforms the image-only baseline on most benchmark tasks, with the most notable gains on gene expression prediction. On HEST-Bench, which requires predicting 50 highly variable genes from H\&E patches, Pearson correlation improves from 0.31 to 0.38. This improvement is partially attributable to multimodal pretraining, though increased exposure to HEST images during continued pretraining may also contribute. \modelname also outperforms the contrastive SEAL-CONCH baseline on this task.

On HEST-Expr (predicting 380 fixed-panel genes from H\&E) and BEAT (predicting 100 Xenium add-on genes from H\&E of the ETOP BEAT-meso data), \modelname improves Pearson correlation over the image-only baseline from 0.26 to 0.36 and from 0.53 to 0.66 respectively. Providing non-target gene expression at inference yields further gains: HEST-Expr reaches 0.36 with 1,000 input genes, and BEAT reaches 0.66 with 380 input genes, demonstrating \modelname's ability to leverage multimodal inputs. A supervised baseline trained directly for the BEAT prediction task \cite{martinelli_learning_2026} — using a similar tokenization but task-specific supervision — outperforms \modelname.

We improve over the image-only base most clearly on regression (HBench, HexprIO, BEATIO) and segmentation tasks (CNP, MNS, TTS), improving 0.01 to 0.08 points over the baseline. On multi-class classification tasks, we observe performance drops on breast cancer benchmarks (BACH, BRACS, BreakHis) and mixed results on prostate tasks: we see improvement on PANDA-Radboud but a loss on Gleason TMA (Tables \ref{tab:benchmark}\&\ref{tab:benchmark2}). These tasks share a common difficulty: distinguishing visually similar cancer subtypes. Continued multimodal pretraining may reorient the encoder toward gene-expression-aligned features, weakening the subtle morphological detail these subtype tasks rely on. Whether this is tissue-specific or task-specific remains open.

In summary we trained \modelname, an early-fusion multimodal transformer that integrates H\&E image and spatial transcriptomics tokens into a shared ViT token sequence, by continuing DINOv2 self-supervised pretraining on paired image and ST data from a vision-only pathology foundation model. \modelname outperforms the image-only baseline and SEAL-CONCH on gene expression prediction tasks HEST-Bench and HEST-Expr, and shows further improvements when transcript inputs are available at inference. Performance on image-only classification tasks is largely preserved, though drops on breast cancer benchmarks suggest a potential trade-off between multimodal representation learning and fine-grained subtype discrimination.
\FloatBarrier

\section*{Impact Statement}
This paper presents work whose goal is to advance the field of Machine Learning. There are many potential societal consequences of our work, none which we feel must be specifically highlighted here.


\bibliography{references}
\bibliographystyle{icml2026}

\newpage
\appendix
\onecolumn
\section{Model details}
\label{app:model_details}

\subsection{Training data}
\begin{itemize}
    \item 7924 \textbf{TCGA}~\cite{tcga_cancer_2025} diagnostic slides, also used in training the baseline backbone, are used to preserve the model's base image processing capabilities.
    \item \textbf{HEST-1k} (v1.1.0): 1,229 WSIs spanning 26 organs and two organisms (human and mouse), comprising 65 Xenium samples and 1,164 spot-based expression samples (Visium and Spatial Transcriptomics) \cite{jaume_hest-1k_2024}. From this we constructed paired image-transcript samples from human Xenium and Visium data at varying image and transcript token resolutions, as well as image-only samples from the WSIs. We follow the official HEST-Bench train/test split, using only the training split during pretraining.
    \item \textbf{ETOP BEAT-meso} data from the ETOP IBCSG Partners Foundation consisting 163 mesothelioma patients with paired H\&E and Xenium measurements using the core Human Immuno-Oncology Panel (380 genes) and an additional custom panel (100 genes) used for both paired H\&E and Xenium training as well as image-only training. We used 122 samples for training and 41 samples for testing.
    
    
\end{itemize}

\subsection{Architecture}
We trained a family of multimodal foundation models that jointly encode tissue morphology and gene expression. The morphology branch is a ViT-B/14 backbone, initialized from a DINOv2 checkpoint pretrained in-house on pathology data and fine-tuned during training. The gene expression branch is a frozen Geneformer encoder (gf-6L-10M-i2048)~\cite{theodoris_transfer_2023}. Its output passes through a \texttt{BatchNorm1d} layer over the embedding dimension, which re-scales each feature to zero mean and unit variance (with a learned affine transform), and is then linearly projected to the morphology embedding dimension. Before fusion, every expression token is additionally rescaled in L2 norm to match its corresponding morphology token (when expression and morphology tokens are aligned one-to-one) or to the mean norm of the morphology patch tokens (when expression tokens are aggregated). The two streams are then fused by token concatenation before being projected by the DINO head. The morphology features were sized as 16$\times$16 tokens per patch.

Variants of the model differ only in how gene expression is aggregated into tokens (Section~\ref{sec:variants}); the rest of the architecture, optimization, and data pipeline is shared across all configurations.

\subsection{Self-supervised framework}
Models were trained with the DINOv2 objective including the iBOT masked-image-modeling auxiliary loss~\cite{oquab_dinov2_2024}. The full criterion is a weighted sum of a DINO loss (weight 1.0), an iBOT loss (weight 1.0), and a KDE regularizer (weight 0.05). Teacher targets were produced with Sinkhorn--Knopp centering and a warmup teacher temperature of 0.07.

The DINO projection head followed the standard three-layer MLP design with input dimension 768, hidden dimension 2048, bottleneck dimension 384, and output dimension 131'072. The last layer of the head was kept frozen for the first three epochs of training. Both the morphology backbone and the projection head were initialized from the same in-house DINOv2 pathology run; the head's prototypes were therefore warm-started rather than learned from scratch.

\subsection{Optimization}
We optimized with AdamW ($\beta = (0.9, 0.999)$) at a base learning rate of $2 \times 10^{-4}$ and an initial weight decay of 0.04. The learning rate followed a cosine schedule with a 10-epoch linear warmup, decaying to 10\% of the base value over 200 epochs. Weight decay was cosine-annealed from 0.04 to 0.2, and the EMA teacher momentum was cosine-annealed from 0.994 to 1.0. Training used BF16 mixed precision, gradient clipping at 1.0, distributed data parallelism (DDP), and a fixed random seed of 42. All models were trained for at most 200 epochs.

\subsection{Data and patch extraction}
Training mixed six data streams via weighted sampling with equal weights for all datasets: BEAT-MESO images (17\%), BEAT-MESO images paired with Xenium transcripts (17\%),HEST-Xenium with paired spatial transcripts (17\%), TCGA image-only (17\%), HEST image-only (17\%), and HEST spot-level expression supplied as cell sentences (17\%). HEST--Xenium patches were filtered to retain at least 100 transcripts and 10 unique genes; HEST cell-sentence inputs were truncated to a maximum of 200 genes per spot.

Image patches were $224 \times 224$ pixels and were sampled at multiple resolutions: 0.5, 1.0, and 2.0\,\textmu m/px for HEST--Xenium, and 0.25, 0.5, 1.0, and 2.0\,\textmu m/px for the TCGA and HEST image-only streams. We applied DINOv2 multi-crop augmentation with two global views ($224 \times 224$) and six local views ($112 \times 112$). iBOT block-wise masking was applied with probability 0.5 and a mask ratio drawn uniformly from $[0.1, 0.5]$.

\subsection{Configuration-specific details}
\label{sec:variants}
The four variants differ in how spatial transcripts are aggregated into expression tokens before fusion. A sliding window with kernel size $k$ and stride $k$ (non-overlapping) is applied over the $224\times224$ patch, yielding $(224/k)^2$ expression tokens per patch (Table~\ref{tab:config-specific-hyperparams}). The 16 token variant produces a coarse spatial grid ($4\times4$), and the 256-token variant produces a $16\times16$ grid that aligns one expression token with each morphology patch token.

To keep optimization comparable across variants, gradient accumulation and per-epoch batch counts were adjusted so that all configurations converged on an effective batch size of 1024 samples and approximately 78 optimization steps per epoch ($\approx 640'000$ training samples per epoch), absorbing differences in the number of GPUs available in each compute environment. The per-GPU micro-batch size was 4 in all configurations.

\begin{table}[ht]
\centering
\caption{Configuration-specific hyperparameters for the four expression token variants.}
\label{tab:config-specific-hyperparams}
\begin{tabular}{lcccc}
\toprule
\textbf{Parameter} &  \textbf{16 tokens} &\textbf{256 tokens} \\
\midrule
Expression aggregation kernel          & $56 \times 56$ & $14 \times 14$ \\
Expression aggregation stride          & 56 & 14 \\
Expression tokens per patch            & 16 ($4 \times 4$) & 256 ($16 \times 16$) \\
GPUs                                   & 2  & 4 \\
Gradient accumulation steps            & 128 & 64 \\
\bottomrule
\end{tabular}
\end{table}

\section{Benchmark tasks details}
\label{app:task_details}

\begin{table*}[t]
  \caption{Benchmark results of all models on all tasks
  Bach \cite{polonia_bach_2019, aresta_bach_2019, araujo_classification_2017, fondon_automatic_2018}, BRACS \cite{brancati_bracs_2022},  BH: BreakHis \cite{spanhol_dataset_2016}, CRC: CRC-100K \cite{kather_100000_2018}, GTMA: Gleason TMA  \cite{arvaniti_automated_2018}, MHIST \cite{wei_petri_2021}, PC: PatchCamelyon \cite{veeling_rotation_2018}, C16: Camelyon16  \cite{ehteshami_bejnordi_diagnostic_2017}, Panda \cite{bulten_artificial_2022}, CNP: CoNSeP \cite{graham_hover-net_2019}, MNS: MoNuSAC \cite{verma_monusac2020_2021}, Hbench: Hest-bench \cite{jaume_hest-1k_2024}, Hexpr: Hest expression prediction data from \cite{jaume_hest-1k_2024}, HexprIO: Hest expression task with image only input, PR: panda radboud \cite{bulten_artificial_2022}, PK: panda karolinska \cite{bulten_artificial_2022}. TTS: tan tock seng \cite{oner_ai-assisted_2022, oner_digital_2022}. Conch+SEAL: \cite{hemker_towards_2026}. Avg: Average over all tasks, excluding Hexpr and BEAT, but including HexprIO and BEATIO}
  \label{tab:benchmark}
  \begin{center}
    \begin{small}
      \begin{sc}
        \begin{tabular}{lllllllllll}
        \toprule
Model                   & BACH & BRACS & BH   & CRC  & GTMA & MHIST & PC   & C16  & Panda & CNP  \\
\midrule
Image only              & 0.88 & 0.64  & 0.82 & 0.96 & 0.77 & 0.77  & 0.91 & 0.79 & 0.63  & 0.61 \\
16 expr tokens          & 0.82 & 0.60  & 0.73 & 0.96 & 0.69 & 0.74  & 0.89 & 0.87 & 0.64  & 0.62 \\
256 expr tokens         & 0.86 & 0.63  & 0.77 & 0.95 & 0.71 & 0.75  & 0.90 & 0.92 & 0.63  & 0.61 \\
Conch + SEAL            & 0.82 & 0.61  & 0.68 & 0.95 & 0.73 & 0.73  & 0.91 & 0.89 & 0.62  & 0.62 \\
\bottomrule
        \end{tabular}
      \end{sc}
    \end{small}
  \end{center}
  \vskip -0.1in
\end{table*}

\begin{table*}[t]
  \caption{}
  \label{tab:benchmark2}
  \begin{center}
    \begin{small}
      \begin{sc}
        \begin{tabular}{lllllllllll}
        \toprule
Model                   & MNS  & HBench & Hexpr & HexprIO & PK   & PR   & TTS   & BEAT & BEATIO & Avg \\
\midrule                                                                                      
Image only              & 0.61 & 0.31   & -     & 0.26    & 0.70 & 0.61 & 0.31  & -    & 0.53   & 0.65 \\
16 expr tokens          & 0.62 & 0.38   & 0.36  & 0.30    & 0.72 & 0.67 & 0.37  & 0.66 & 0.58   & 0.66 \\
256 expr tokens         & 0.64 & 0.38   & 0.37  & 0.30    & 0.72 & 0.67 & 0.38  & 0.67 & 0.59   & 0.67 \\
Conch + SEAL            & 0.63 & 0.35   &  -    & 0.28    & 0.71 & 0.66 & 0.37  &  -   & -      & - \\
\bottomrule
        \end{tabular}
      \end{sc}
    \end{small}
  \end{center}
  \vskip -0.1in
\end{table*}

\subsection{Classification tasks}
\subsubsection{BACH}
This dataset consists of 400 (268 train / 132 val) WSIs of breast cancer, split by patient ID to avoid leakage \cite{polonia_bach_2019, aresta_bach_2019, araujo_classification_2017, fondon_automatic_2018}. The task is patch-level multiclass classification task, predicting 4 classes: normal, benign, in situ, invasive, measured by multiclass accuracy. We do linear probing using nn.Linear(out=4), CrossEntropyLoss, AdamW lr=3e-4, batch size 256, max 12,500 steps, early stopping patience 1250 on val/MulticlassAccuracy.

\subsubsection{BRACS}
Data contains 3,657 train / 312 val / 570 test ROI patches of breast lesions \cite{brancati_bracs_2022}. The task is patch-level multiclass classification task, predicting 7 subtypes: Normal, Pathological Benign, Usual Ductal Hyperplasia, Flat Epithelial Atypia, Atypical Ductal Hyperplasia, Ductal Carcinoma In Situ, Invasive Carcinoma, measured by multiclass accuracy We do linear probing with nn.Linear (out=7), CrossEntropyLoss, AdamW lr=3e-4, batch size 256, max 12500 steps, early stopping patience 167 on val/MulticlassAccuracy.

\subsubsection{BreakHis}
Data has 1,132 train / 339 val WSIs of breast cancer patients, split by patient ID (14 patients held out for val) to prevent data leakage \cite{spanhol_dataset_2016}. The task is patch-level multiclass classification task, predicting 4 classes: Tubular Adenoma (TA), Mucinous Carcinoma (MC), Fibroadenoma (F), Ductal Carcinoma (DC), measured by multiclass accuracy We performed linear probing with nn.Linear (out=4), CrossEntropyLoss, AdamW lr=3e-4, batch size 256, max 12500 steps, early stopping patience 500 on val/MulticlassAccuracy. 

\subsubsection{CRC-100K}
The dataset contains 100'000 train / 7180 val patches, split by patient ID 	\cite{kather_100000_2018}. The task is patch-level multiclass classification task, predicting 9 classes: Adipose (ADI), Background (BACK), Debris (DEB), Lymphocytes (LYM), Mucus (MUC), Smooth Muscle (MUS), Normal Mucosa (NORM), Cancer-associated Stroma (STR), Colorectal Adenocarcinoma (TUM), measured by multiclass accuracy We do linear probing with nn.Linear (out=9), CrossEntropyLoss, AdamW lr=3e-4, batch size 256, max 12500 steps, early stopping patience 7 on val/MulticlassAccuracy.

\subsubsection{Gleason TMA}
The data consists of 15303 train / 2482 val / 4967 test patches of prostate cancer, split by tissue micro-array ID \cite{arvaniti_automated_2018}. The task is multiclass patch classification of higher level glandular patterns in 4 classes: Benign, Gleason 3, Gleason 4, Gleason 5, measured by multiclass accuracy. We do linear probing with: nn.Linear (out=4), CrossEntropyLoss, AdamW lr=3e-4, batch size 256, max 12500 steps, early stopping patience 42 on val/MulticlassAccuracy.

\subsubsection{MHIST}
This dataset consists of 3,152 (2,175 train / 977 test) colorectal polyp image patches, labeled by majority vote of 7 pathologists \cite{wei_petri_2021}. The task is a patch-level binary classification task, distinguishing hyperplastic polyps (HP) from sessile serrated adenomas (SSA), measured by binary balanced accuracy. We do linear probing using nn.Linear(out=1), BCEWithLogitsLoss, AdamW lr=3e-4, batch size 256, max 12500 steps, early stopping patience 278 on val/BinaryBalancedAccuracy.

\subsubsection{Patch-camelyon}
This dataset consists of 327,680 (262,144 train / 32,768 val / 32,768 test) 96x96 px patches extracted from Camelyon16 lymph node WSIs \cite{veeling_rotation_2018}. The task is a patch-level binary classification task, detecting tumor metastasis vs. normal tissue, measured by binary balanced accuracy. We do linear probing using nn.Linear(out=1), BCEWithLogitsLoss, AdamW lr=3e-4, batch size 256, max 12,500 steps, early stopping patience 3 on val/BinaryBalancedAccuracy.

\subsubsection{Camelyon16}
This dataset consists of 399 (216 train / 54 val / 129 test) lymph node WSIs, with validation slides matching those used to derive PatchCamelyon \cite{ehteshami_bejnordi_diagnostic_2017}. The task is a slide-level binary classification task, detecting tumor metastasis vs. normal tissue, measured by binary balanced accuracy. We use a two-stage offline pipeline: first extract up to 1,000 patch embeddings per slide (224x224 at 0.25 mpp), then train ABMIL (projected dim 128, out=1), BCEWithLogitsLoss, AdamW lr=1e-3, batch size 32, max 100 epochs, early stopping patience 20 on val/BinaryBalancedAccuracy, averaged over 20 runs.

\subsubsection{Panda}
This dataset consists of ~10,000 prostate cancer WSIs from the Kaggle PANDA challenge, split 70/15/15 stratified by grade, with noisy labels filtered out \cite{bulten_artificial_2022}. The task is a slide-level multiclass classification task, predicting 6 ISUP grades (0–5), measured by multiclass accuracy. We use a two-stage offline pipeline: first extract up to 200 patch embeddings per slide (224x224 at 0.5 mpp), then train ABMIL (projected dim 128, out=6), CrossEntropyLoss, AdamW lr=1e-3, batch size 32, max 100 epochs, early stopping patience 20 on val/MulticlassAccuracy, averaged over 20 runs.

\subsubsection{Panda radboud}
This dataset is derived from the Radboud subset of the PANDA prostate cancer challenge \cite{bulten_artificial_2022}. Patches of 256x256 at 0.25 mpp are extracted from WSIs and filtered to contain exactly one Gleason grade covering at least 30\% of the patch area. The task is a patch-level multiclass classification task, predicting 5 Gleason grades (stroma, benign epithelium, Gleason 3, 4, 5), with up to 25,000 balanced patches sampled (capped at 200 patches/WSI, 20 patches/WSI/class). Labels come from pixel-level segmentation masks where prostate glands are individually annotated. We use an offline two-stage pipeline with 5-fold cross-validation: first extract and cache patch embeddings with a frozen encoder, then do linear probing using nn.Linear(out=5), CrossEntropyLoss, AdamW lr=3e-4, batch size 256, max 12,500 steps, early stopping patience 278 on val/MulticlassAccuracy.
\subsubsection{Panda karolinska}
This dataset is derived from the Karolinska subset of the PANDA prostate cancer challenge \cite{bulten_artificial_2022}. Patches of 256x256 at 0.25 mpp are extracted from WSIs and filtered to contain exactly one tissue class covering at least 30\% of the patch area. The task is a patch-level binary classification task, predicting 2 classes (benign tissue vs. cancerous tissue), with up to 1,000,000 balanced patches sampled (capped at 350 patches/WSI, 200 patches/WSI/class). Karolinska provides coarser region-level annotations compared to Radboud's gland-level labels. We use an offline two-stage pipeline with 5-fold cross-validation: first extract and cache patch embeddings with a frozen encoder, then do linear probing using nn.Linear(out=1), BCEWithLogitsLoss, AdamW lr=3e-4, batch size 256, max 12,500 steps, early stopping patience 278 on val/BinaryBalancedAccuracy.

\subsection{Segmentation tasks}
\subsubsection{CoNSeP (Colorectal Nuclear Segmentation and Phenotypes)}
This task includes 41 H\&E-stained tiles (27 train / 14 val images, 1000x1000px) from 16 colorectal adenocarcinoma patients, containing 24,319 annotated nuclei \cite{graham_hover-net_2019}. The goal is to semantically segment into five classes: background, other, inflammatory, epithelial, spindle-shaped. 
We used patch\_size 224x224 and batch\_size=64. Light-weight decoder probe is trained with early stopping with the MonaiDiceScore, best decoder probe is used for the test set.

\subsubsection{MoNuSAC (Multi-organ Nuclei Segmentation and Classification)}
Dataset consists of ~46,000 annotated nuclei from TCGA WSIs across multiple organs at 40x magnification and 209 train/ 85 test images	\cite{verma_monusac2020_2021}. Semantic segmentation task into 5 classes: background, epithelial, lymphocyte, neutrophil, macrophage. Light-weight decoder probe is trained with early stopping with the MonaiDiceScore, best decoder probe is used for the test set.

\subsubsection{Tan tock seng}
This dataset is from Tan Tock Seng Hospital, Singapore \cite{oner_ai-assisted_2022, oner_digital_2022}. H\&E-stained WSIs from 99 prostate cancer patients (40 prostatectomy + 59 core needle biopsy specimens) were scanned at 40x magnification (0.25 µm/px) using an Aperio AT2 scanner. Pre-extracted patches of 512x512 at the 50 µm resolution level are used from the gland classification subset (46 WSIs), with patient-level train/val/test splits. The task is a patch-level semantic segmentation task predicting 9 classes (background, Gleason pattern 3, Gleason pattern 4, benign, tangential benign, tangential malignant, unknown, PIN, artefact). We use an offline two-stage pipeline with 5 runs: first extract and cache patch embeddings with a frozen encoder (resized to 224x224, ImageNet normalization), then train a ConvDecoderMS segmentation decoder using DiceLoss (softmax, batch mode), AdamW lr=2e-3, batch size 64, max 2,000 steps, early stopping patience 200 on val/MonaiDiceScore. The primary metric is classwise MonaiDiceScore (including background).
\subsection{Regression tasks}
\subsubsection{HEST-bench}
The hest benchmark task is an image to gene expression prediction tasks, made available by \cite{jaume_hest-1k_2024} with 10 cancer types and Visium and Xenium samples. The image patches, cross-validation folds, and 50 target highly variable genes per cancer type are given. The task is evaluated by Pearson correlation. Below we report the task-specific performance per model.
\begin{table*}[t]
  \caption{HEST-bench benchmark results for all models, shown in the average Pearson Correlation coefficient}
  \label{tab:hest-bench}
  \begin{center}
    \begin{small}
      \begin{sc}
        \begin{tabular}{lccccccccccc}
        \toprule
Model                   & CCRCC & COAD  & HCC   & IDC   & LUNG  & LYMPH\_IDC & PAAD  & PRAD  & READ  & SKCM  \\
\midrule
Image only              & 0.096 & 0.270 & 0.083 & 0.406 & 0.513 & 0.298      & 0.350 & 0.309 & 0.171 & 0.610 \\
16 expr tokens          & 0.192 & 0.334 & 0.082 & 0.439 & 0.615 & 0.370      & 0.431 & 0.422 & 0.245 & 0.707 \\
256 expr tokens         & 0.198 & 0.348 & 0.091 & 0.438 & 0.623 & 0.362      & 0.4237 & 0.399 & 0.248 & 0.689 \\
Conch + SEAL            & 0.182 & 0.288 & 0.047 & 0.425 & 0.553 & 0.340      & 0.385 & 0.361 & 0.228 & 0.652 \\
\bottomrule
        \end{tabular}
      \end{sc}
    \end{small}
  \end{center}
  \vskip -0.1in
\end{table*}

\subsubsection{HEST-expr}
This benchmark evaluates multimodal reasoning by predicting a target gene expression panel from H\&E images and an input gene expression panel. It is inspired by the practical challenge of limited Xenium panels, where measured genes may not overlap across samples. Specifically, we predict the v1-human-immuno-oncology panel (380 genes) from 10X Genomics, chosen for its relevance to cancer-related tasks, from H\&E patches and 1,000 stable input genes.

To select the 1,000 input genes, we used all Visium samples in HEST-Bench, which are held out during training. We first applied spot- and sample-level quality control (QC): spots were excluded if out-of-tissue, more than 5 MAD below the per-sample median on log1p(total\_counts) or log1p(n\_genes\_by\_counts), or if pct\_counts\_mito exceeded both 5 MAD above the median and 20\%. Samples were excluded if fewer than 1,000 spots passed QC, median counts fell below 1,000, or the fraction of spots passing QC was below 0.75 (one sample was filtered). We then ranked genes within each sample by detection rate (fraction of spots expressing gene $G$) and mean log1p normalized expression, averaged ranks across samples, and computed a composite rank weighting both criteria equally. The final 1,000 input genes were selected as those with the highest composite rank, a minimum detection rate of 0.05 in every sample, and no overlap with the target v1-human-immuno-oncology panel. Note that while QC is applied during gene selection, all spots and samples are included in the benchmark evaluation itself.

\subsection{BEAT}
This task evaluates the ability of multimodal (image + expression) encoders to predict unmeasured gene expression from a held-out set of 100 add-on genes ($G_\text{add}$) that are not part of the core Xenium Human Immuno-Oncology Panel ($G_\text{core}$ = 380 genes). The private dataset consists of 163 mesothelioma patients with paired H\&E and Xenium measurements, of which 41 are in the test split. For evaluation, patches of 224x224 at 0.5 mpp are extracted from the test split, with min 1,000 transcripts and min 40\% foreground per patch. The add-on genes are aggregated per patch and log1p-transformed. The task uses Ridge regression with 4-fold group-stratified cross-validation (grouped by patient ID) to predict non-panel gene expression from the frozen encoder's CLS token embedding. We report the Pearson correlation coefficient per gene and averaged over all validation patches. 

\subsection{BEATIO}
Same task and evaluation protocol as above, but using only the morphology (image) encoder backbone without expression token input. This isolates the image encoder's contribution to gene expression prediction and serves as a unimodal baseline, measuring how much predictive signal for non-panel genes can be inferred from tissue morphology alone versus the multimodal representation.

\end{document}